\documentclass[letterpaper]{article} 
\usepackage{aaai2026}  
\usepackage{times}  
\usepackage{helvet}  
\usepackage{courier}  
\usepackage[hyphens]{url}  
\usepackage{graphicx} 
\usepackage{natbib}  
\usepackage{caption} 
\usepackage{algorithm}
\usepackage{algorithmic}
\usepackage{amsmath}
\usepackage{booktabs} 
\usepackage{enumitem} 
\usepackage{amsfonts}
\usepackage{csquotes}
\usepackage{booktabs}
\usepackage{graphicx} 
\usepackage{multirow}
\usepackage{graphicx}
\usepackage{subcaption}
\usepackage{placeins}
\usepackage[table]{xcolor}
\usepackage{bm}

\usepackage{newfloat}
\usepackage{listings}
\DeclareCaptionStyle{ruled}{labelfont=normalfont,labelsep=colon,strut=off} 
\floatstyle{ruled}
\newfloat{listing}{tb}{lst}{}
\floatname{listing}{Listing}
\title{JUSTICE: Judicial Unified Synthesis Through Intermediate Conclusion Emulation for Automated Judgment Document Generation}

\author {
    Binglin Wu\textsuperscript{\rm {1,}}\thanks{Equal contribution.},
    Yingyi Zhang\textsuperscript{\rm {1,2}}\footnotemark[1],
    Xianneng Li\textsuperscript{\rm {1,}}\thanks{Corresponding author.}
}
\affiliations {
    \textsuperscript{\rm 1}Dalian University of Technology, 
    \textsuperscript{\rm 2}City University of Hong Kong
}

\usepackage{bibentry}

\begin{document}
\nocopyright
\maketitle

\begin{abstract}

Automated judgment document generation is a significant yet challenging legal AI task. As the conclusive written instrument issued by a court, a judgment document embodies complex legal reasoning. However, existing methods often oversimplify this complex process, particularly by omitting the ``Pre-Judge'' phase—a crucial step where human judges form a preliminary conclusion.  This omission leads to two core challenges: 1) the ineffective acquisition of foundational judicial elements, and 2) the inadequate modeling of the Pre-Judge process, which collectively undermine the final document's legal soundness. To address these challenges, we propose \textit{\textbf{J}udicial \textbf{U}nified \textbf{S}ynthesis \textbf{T}hrough \textbf{I}ntermediate \textbf{C}onclusion \textbf{E}mulation} (JUSTICE), a novel framework that emulates the ``Search $\rightarrow$ Pre-Judge $\rightarrow$ Write'' cognitive workflow of human judges. Specifically, it introduces the Pre-Judge stage through three dedicated components: Referential Judicial Element Retriever (RJER), Intermediate Conclusion Emulator (ICE), and Judicial Unified Synthesizer (JUS). RJER first retrieves legal articles and a precedent case to establish a referential foundation. ICE then operationalizes the Pre-Judge phase by generating a verifiable intermediate conclusion. Finally, JUS synthesizes these inputs to craft the final judgment. Experiments on both an in-domain legal benchmark and an out-of-distribution dataset show that JUSTICE significantly outperforms strong baselines, with substantial gains in legal accuracy, including a 4.6\% improvement in prison term prediction. Our findings underscore the importance of explicitly modeling the Pre-Judge process to enhance the legal coherence and accuracy of generated judgment documents.

\end{abstract}

\section{Introduction}

\begin{figure}[h]
    \centering
    \includegraphics[width=0.9\columnwidth]{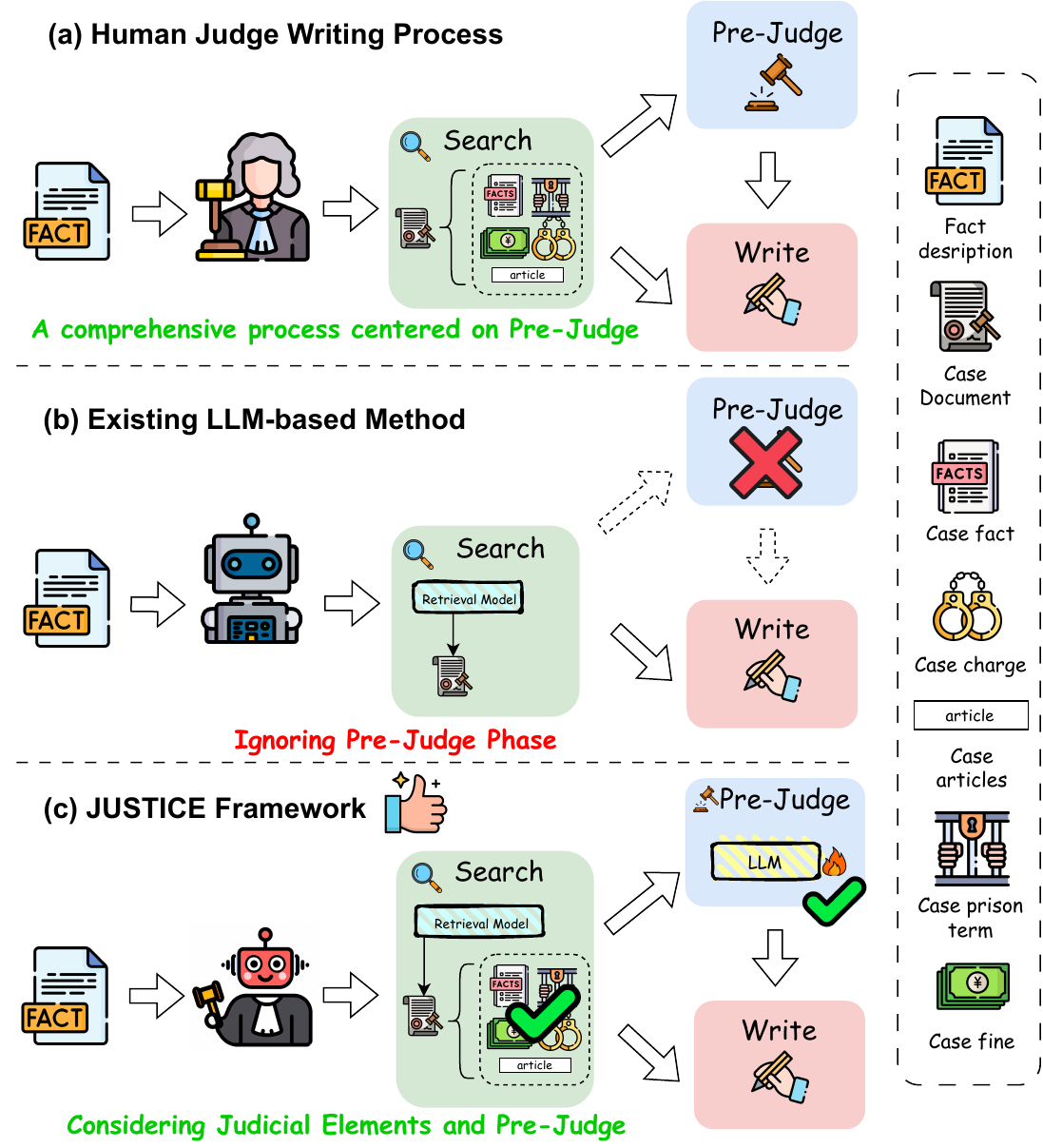} 
    \caption{Illustration of the challenges of the existing Judgment Document Generation method.}
    \vspace{-10pt}
    \label{fig:intro}
\end{figure}

In recent years, the rapid advancement of Large Language Models (LLMs)~\cite{cui2023chatlaw,yang2023baichuan,yang2025qwen2} has led to their widespread application in the legal domain for tasks such as similar case retrieval~\cite{li2025delta,gao2024enhancing} and legal question answering~\cite{hu2025fine,louis2024interpretable}. Among the numerous legal AI tasks, automated judgment document generation stands out as a significant yet challenging problem. As an essential part of the judicial process, a judgment document is the conclusive written instrument issued by a court after adjudicating a case, embodying complex legal reasoning~\cite{kumar2013finding,zander2015law}. In practice, the writing of these documents by human judges is a time-consuming and labor-intensive endeavor. As illustrated in Figure~\ref{fig:intro}(a), this process typically involves three stages: 1) a Search stage to consult extensive legal materials, including applicable statutes and reference precedents; 2) a Pre-Judge stage to integrate these materials with case facts and form a preliminary conclusion; and 3) a Write stage to compose the final judgment that articulates the underlying legal reasoning.

The task of judgment document generation requires a system to automatically create a complete legal document from a case's fact description, encompassing sections such as litigant information, findings of fact, adjudication reasoning, and the final judgment result~\cite{su2025judge}. Although the advent of LLMs has simplified text generation, producing a Chinese judgment document demands more than mere textual fluency; it necessitates complex legal reasoning, precise citation of legal articles, and the integration of key case facts. As illustrated in Figure~\ref{fig:intro}(b), existing LLM-based methods typically follow a simplified pipeline: after receiving the case facts, they directly proceed to information retrieval and then enter the writing phase. Such a simplified workflow reveals a significant gap when compared to the actual case-handling process of human judges. This oversight is critical, as it prevents the establishment of a sound legal basis for the judgment, directly undermining the final document's accuracy and legal coherence.

In the pursuit of automated judgment generation, LLMs are currently unable to replicate the core role of a human judge.  The key bottleneck is the absence of a Pre-Judge phase, a critical step that human judges undertake before delivering a final ruling.  Specifically, this presents two primary challenges: \textbf{1) Effective Acquisition of Judicial Elements for Pre-Judge. } The initial step of Pre-Judge requires the model to emulate a human judge by retrieving core adjudicative elements from vast data to form a basis for its conclusion. These elements include relevant articles of law and the judgments of historically similar cases. The absence or low quality of these retrieved elements can severely compromise the accuracy of the subsequent Pre-Judge conclusion. \textbf{2) Effective Modeling of the Pre-Judge Process.}  After acquiring the necessary judicial elements, the model must process and weigh them to formulate a preliminary judgment.  Without an effective model for this process, the LLM cannot perform judge-like analogical reasoning to arrive at a high-quality preliminary conclusion, even when provided with the correct articles and case precedents. These challenges raise a critical question: \textit{How can we design a judgment document generation framework that effectively and agentically incorporates the Pre-Judge process to produce documents excelling in both judgmental accuracy and textual fluency?}

To address these challenges, as shown in Figure~\ref{fig:intro}(c), we introduce the JUSTICE framework, which emulates the ``Search $\rightarrow$ Pre-Judge $\rightarrow$ Write'' cognitive workflow of human judges through three dedicated components: Referential Judicial Element Retriever (RJER), Intermediate Conclusion Emulator (ICE), and Judicial Unified Synthesizer (JUS). \textbf{RJER: Establishing a Referential Foundation.} To tackle the element acquisition challenge, RJER first retrieves both relevant legal articles and a complete precedent case with its core judicial elements.
\textbf{ICE: Operationalizing the Pre-Judge Phase.} To address the modeling challenge, ICE then simulates a judge's preliminary assessment to generate a verifiable, structured intermediate conclusion from the retrieved elements.
\textbf{JUS: Crafting a Coherent Final Document.} Finally, JUS synthesizes the case facts and the intermediate conclusion under the guidance of the precedent case to generate a final judgment that is both legally coherent and textually fluent.

Our contributions are as follows:
\begin{itemize}[leftmargin=*]
    \item We are the first to propose a novel framework that emulates the complete ``Search $\rightarrow$ Pre-Judge $\rightarrow$ Write''  cognitive workflow of human judges.
    \item We introduce three core components: RJER retrieves judicial elements in parallel; ICE forms a structured intermediate conclusion using an LLM; and JUS synthesizes the final document with another LLM.
    \item Extensive experiments show that JUSTICE significantly outperforms strong baselines, validating the critical role of the Pre-Judge stage in generating judgment documents.

\end{itemize}

\section{Problem Definition}
\label{sec:problem_definition}

The task of Judgment Document Generation is defined as a mapping from a case's fact description, represented by the variable $F$, to a complete and legally valid judgment document $D$. The document $D$ must be composed of several sections, including the heading, fact description, judicial reasoning, and the final judgment result.

Inspired by the cognitive process of human judges, we decompose the overall generation function $\Psi$ into a three-stage pipeline that emulates the ``Search $\rightarrow$ Pre-Judge $\rightarrow$ Write'' workflow.

First, the Search function $\Psi_{\text{Search}}$ takes the case facts $F$ to retrieve a composite set of referential judicial elements, which we denote as $\mathcal{E}_{\text{ref}}$. This set specifically consists of three parts: the extracted key elements from a precedent case $\mathcal{E}_{\text{case}}$, supplementary legal articles $\mathcal{A}_{\text{ext}}$, and the full document of the precedent case $\mathcal{C}_{\text{doc}}$.
\begin{equation}
    \mathcal{E}_{\text{ref}} = \{ \mathcal{E}_{\text{case}}, \mathcal{A}_{\text{ext}}, \mathcal{C}_{\text{doc}} \} = \Psi_{\text{Search}}(F) \label{eq:search}
\end{equation}

Next, the Pre-Judge function $\Psi_{\text{Pre-Judge}}$ processes the facts $F$ and a specific subset of the retrieved elements—$\mathcal{E}_{\text{case}}$ and $\mathcal{A}_{\text{ext}}$—to produce a structured intermediate conclusion $J_{\text{pre}}$.
\begin{equation}
    J_{\text{pre}} = \Psi_{\text{Pre-Judge}}(F, \mathcal{E}_{\text{case}}, \mathcal{A}_{\text{ext}}) \label{eq:pre-judge}
\end{equation}

Finally, the Write function $\Psi_{\text{Write}}$ synthesizes the facts $F$, the intermediate conclusion $J_{\text{pre}}$, and the full precedent document $\mathcal{C}_{\text{doc}}$ to craft the final judgment document $D$.
\begin{equation}
    D = \Psi_{\text{Write}}(F, J_{\text{pre}}, \mathcal{C}_{\text{doc}}) \label{eq:write}
\end{equation}

\begin{figure*}[th]
    \centering
    \includegraphics[width=0.98\linewidth]{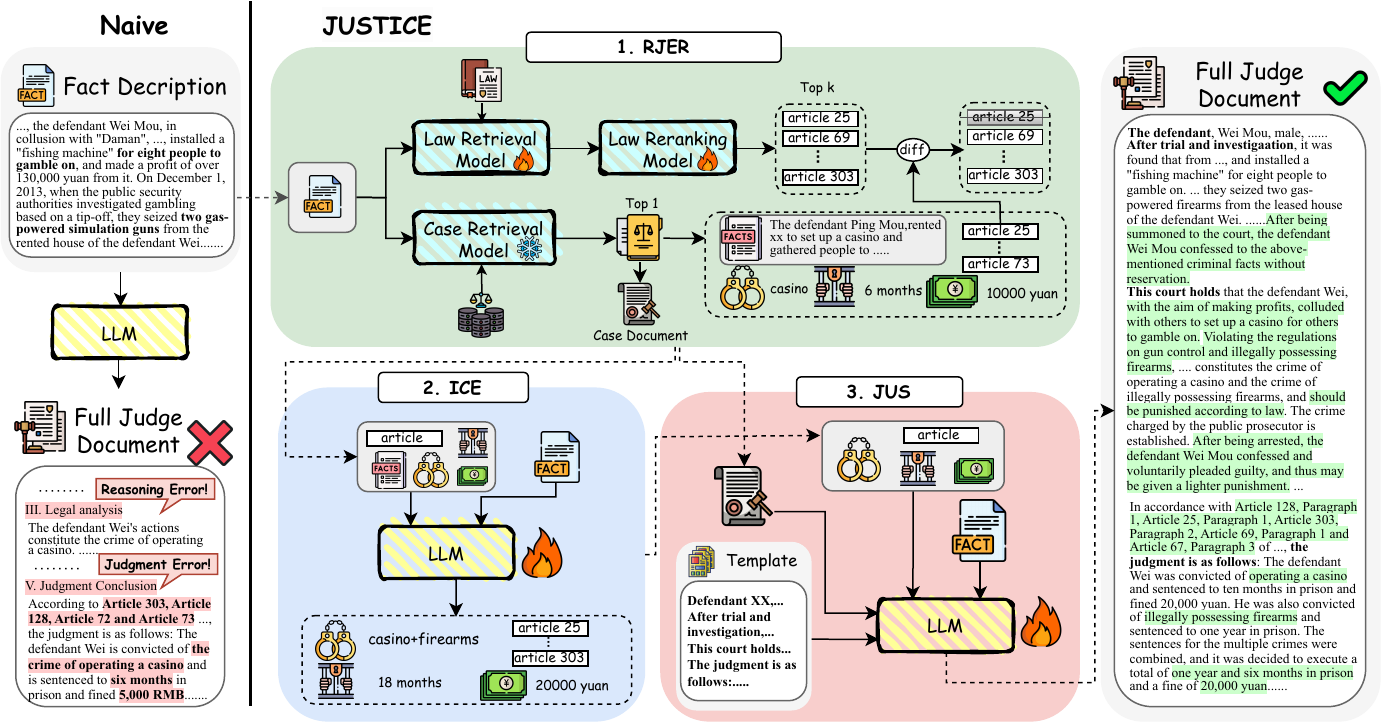}
    \caption{Main framework of JUSTICE.}
    \vspace{-10pt}
    \label{fig:main_framework}
\end{figure*}

\section{JUSTICE Framework}

We propose the JUSTICE framework to emulate the judgment process of human judges, addressing the reliability challenges arising from the omission of the Pre-Judge phase. As illustrated in Figure~\ref{fig:main_framework}, there are three core components: RJER, which retrieves referential judicial elements from laws and past cases; ICE, which forms an intermediate conclusion by emulating a Pre-Judge phase; and JUS, which synthesizes these inputs to craft the final judgment document.

\subsection{Reference Judicial Element Retriever (RJER)}
\label{ssec:RJER}
To address the challenge of ineffective acquisition of referential judicial elements, the RJER component is designed to establish a referential foundation. 

\paragraph{Law Retrieval Model}

First, an efficient bi-encoder model $\Phi_{\text{ret}}$ is employed to retrieve candidate legal articles from the corpus $D_{\text{law}}$. The model is trained using a contrastive objective~\cite{su2025judge} to maximize the relevance score $S_{\text{ret}}(F, A^{+})$ for a positive article $A^{+}$ over scores for negative articles $A^{-}\in\mathcal{N}$:
\begin{equation}
    \mathcal{L}_{\text{ret}} = -\log\frac{\exp(S_{\text{ret}}(F, A^{+}))}{\sum_{A_j \in \{A^{+}\} \cup \mathcal{N}} \exp(S_{\text{ret}}(F, A_j))} 
\end{equation}
where the score $S_{\text{ret}}(F, A) = \Phi_{\text{ret}}(F) \cdot \Phi_{\text{ret}}(A)$  is the dot product of the embeddings. Based on these scores, the model retrieves an initial set of top-$k_1$ candidate articles $\mathcal{A}_{\text{cand}}$:
\begin{equation}
    \mathcal{A}_{\text{cand}} = \underset{A_i \in D_{\text{law}}}{\text{Top-K}} \left( S_{\text{ret}}(F, A_i) \right)
\end{equation}

\paragraph{Law Reranking Model}

To refine the initial candidates $\mathcal{A}_{\text{cand}}$, a more powerful cross-encoder, $\Phi_{\text{rerank}}$, is employed. It is trained with a Localized Contrastive Estimation (LCE) objective~\cite{gao2021rethink} to better distinguish hard negatives:
\begin{equation}
    \mathcal{L}_{\text{rerank}} = -\log\frac{\exp(S_{\text{rerank}}(F, A^{+}))}{\sum_{A_j \in \mathcal{G}_q} \exp(S_{\text{rerank}}(F, A_j))} 
\end{equation}
where $\mathcal{G}_q$ contains one positive and several hard negative articles sampled from $\mathcal{A}_{\text{cand}}$. After reranking, the model selects the top-$k_2$ articles from $\mathcal{A}_{\text{cand}}$ to form the final set of retrieved articles $\mathcal{A}_{\text{ret}}$:
\begin{equation}
    \mathcal{A}_{\text{ret}} = \underset{A_j \in \mathcal{A}_{\text{cand}}}{\text{Top-K}} \left( S_{\text{rerank}}(F, A_j) \right)
\end{equation}

\paragraph{Case Retrieval Model} 

In parallel with article retrieval, a structure-aware model like SAILER~\cite{li2023sailer} is used to retrieve the single most relevant precedent case $\mathcal{C}_{\text{doc}}$ from the case repository $D_{\text{case}}$:
\begin{equation}
    \mathcal{C}_{\text{doc}} = \underset{C \in D_{\text{case}}}{\text{Top-1}} \ \left(\text{SAILER}(F, C)\right)
\end{equation}
The key judgment elements are then extracted from this document using a rule-based method, which we formalize as an extraction function $\Psi_{\text{Extract}}$:
\begin{equation}
    \mathcal{E}_{\text{case}} = \Psi_{\text{Extract}}(\mathcal{C}_{\text{doc}})
\end{equation}
where the extracted elements are denoted as a structured set $\mathcal{E}_{\text{case}} = \{c_{\text{fact}},c_{\text{charge}}, \mathcal{A}_{\text{case}}, c_{\text{term}}, c_{\text{fine}}\}$. These elements provide a rich, contextual example for the subsequent modules.

\paragraph{Element Set Combination}
Finally, the outputs from the two retrieval streams are combined. To provide supplementary legal articles, the external article set $\mathcal{A}_{\text{ext}}$ is defined as the set difference between all retrieved articles $\mathcal{A}_{\text{ret}}$ and those from the precedent case $\mathcal{A}_{\text{case}}$:
\begin{equation}
    \mathcal{A}_{\text{ext}} = \{ a \mid a \in \mathcal{A}_{\text{ret}} \land a \notin \mathcal{A}_{\text{case}} \}
\end{equation}
The complete set of referential elements $\mathcal{E}_{\text{ref}}$ is then constructed as:
\begin{equation}
    \mathcal{E}_{\text{ref}} = \{\mathcal{E}_{\text{case}}, \mathcal{A}_{\text{ext}}, \mathcal{C}_{\text{doc}}\}
\end{equation}
This complete set provides the necessary inputs for both the ICE and JUS modules.

\subsection{Intermediate Conclusion Emulator (ICE)}
\label{ssec:ICE}
To address the challenge of inadequately modeling the Pre-Judge process, the ICE component serves as the framework's core predictive engine. It generates a verifiable, structured intermediate conclusion that simulates a judge's preliminary assessment and grounds the final judgment.

\paragraph{Prediction as Analogy}
The core task of ICE is to generate a structured tuple $J_{\text{pre}}$, representing the intermediate conclusion. This tuple consists of four predicted elements: the applicable articles $p_{\text{articles}}$, the charge $p_{\text{charge}}$, the prison term $p_{\text{term}}$, and the fine $p_{\text{fine}}$. Conceptually, this process emulates a judge's analogical reasoning, where the model must infer the outcome for the current case facts $F$ based on the precedent's own facts and key elements $\mathcal{E}_{\text{case}}$ and supplementary articles $\mathcal{A}_{\text{ext}}$. We formalize this analogy function $\Psi_{\text{analogize}}$ as:

\begin{equation}
    \label{eq:analogize}
    \overbrace{
    \begin{pmatrix}
        p_{\text{articles}} \\
        p_{\text{charge}} \\
        p_{\text{term}} \\
        p_{\text{fine}}
    \end{pmatrix}
    }^{\text{\small Prediction }J_{\text{pre}}}
    = \Psi_{\text{analogize}} \left( F, \underbrace{
    \begin{pmatrix}
        c_{\text{fact}} \\
        \mathcal{A}_{\text{case}} \\
        c_{\text{charge}} \\
        c_{\text{term}} \\
        c_{\text{fine}}
    \end{pmatrix}
    }_{\text{\small Precedent } \mathcal{E}_{\text{case}}}, \mathcal{A}_{\text{ext}} \right)
\end{equation}

\paragraph{Model Objective}

To implement this, we first construct a structured prompt $\mathcal{P}_{\text{ICE}}$ by concatenating the inputs using a template $T_{\text{ICE}}$:

\begin{equation}
    \mathcal{P}_{\text{ICE}} = T_{\text{ICE}}(F \oplus \mathcal{E}_{\text{case}} \oplus \mathcal{A}_{\text{ext}})
\end{equation}

The process is then formalized as a conditional generation task where the model $\Phi_{\text{LLM}}$ learns to produce the structured tuple $J_{\text{pre}}$ based on the information provided in the prompt. Specifically, the model is fine-tuned to find the optimal tuple $J_{\text{pre}}^*$ by maximizing the probability of the ground-truth tuple given the prompt:
\begin{equation}
    J_{\text{pre}}^* = \operatorname*{arg\,max}_{J_{\text{pre}}} P(J_{\text{pre}} | \Phi_{\text{LLM}}(\mathcal{P}_{\text{ICE}}))
\end{equation}

\subsection{Judicial Unified Synthesizer (JUS)}
\label{ssec:JUS}

The JUS component acts as a writer to generate the complete judgment document. It first synthesizes the case facts $F$, the intermediate conclusion $J_{\text{pre}}$, the precedent document $\mathcal{C}_{\text{doc}}$, and a document template $T_{\text{format}}$ into a single prompt, $\mathcal{P}_{\text{JUS}}$:
\begin{equation}
    \mathcal{P}_{\text{JUS}} = T_{\text{JUS}}(F \oplus J_{\text{pre}} \oplus \mathcal{C}_{\text{doc}} \oplus T_{\text{format}})
\end{equation}

A separate LLM $\Phi'_{\text{LLM}}$ is then fine-tuned on this prompt to generate the final document $D = \{w_1, \dots, w_T\}$. The model is trained using a standard language modeling objective to minimize the negative log-likelihood:
\begin{equation}
    \mathcal{L}_{\text{JUS}} = -\sum_{t=1}^{T} \log P(w_t | w_{<t}, \mathcal{P}_{\text{JUS}})
\end{equation}

\newcommand{\Hline}{\noalign{\hrule height 1.2pt}}

\begin{table*}[t!]
\setlength{\tabcolsep}{2mm} 
\centering
\resizebox{0.9\textwidth}{!}{
\begin{tabular}{lcccccccccccc}
\Hline 
\multirow{2}{*}{\textbf{Models}} & 
\multicolumn{2}{c}{\textbf{Penalty Acc}} & 
\multicolumn{3}{c}{\textbf{Convicting Acc}} & 
\multicolumn{3}{c}{\textbf{Referencing Acc}} & 
\multicolumn{2}{c}{\textbf{Reasoning Section}} & 
\multicolumn{2}{c}{\textbf{Judgment Section}} \\ 
\cmidrule(lr){2-3} \cmidrule(lr){4-6} \cmidrule(lr){7-9} \cmidrule(lr){10-11} \cmidrule(l){12-13} 
 & Prison & Fine & Recall & Prec & F1 & Recall & Prec & F1 & METH & BERTS & METH & BERTS  \\
\Hline
\multicolumn{10}{l}{\emph{\textcolor{gray!120}{JuDGE: Traditional CVG Methods}}} \\ 
C3VG(Bart)  & - & - &  - & - & - & - & - & - & 0.2188 & 0.7285 & - & -\\
$\mathrm{EGG_{free}}$  & - & - &  - & - & - & - & - & - & 0.2621 & 0.6991 & - & -\\
EGG  & - & - &  - & - & - & - & - & - & 0.2691 & 0.7188 & - & -\\
\hline
\multicolumn{10}{l}{\emph{\textcolor{gray!120}{JuDGE: Traditional LJP Methods}}} \\ 
TopJudge* & 0.3891 & - & 0.2777 & 0.2854 & 0.2641 & 0.1107 & 0.4531 & 0.1779 & - & - & - & -\\
MPBFN* & 0.3200 & - & 0.3890 & 0.3687 & 0.3583 & 0.1218 & 0.4950 & 0.1955 & - & - & - & - \\
NeurJudge* & 0.3440 & - & 0.5814 & 0.5684 & 0.5633 & 0.1049 & 0.4351 & 0.1691 & - & - & - & - \\
\hline
\multicolumn{10}{l}{\emph{\textcolor{gray!120}{JuDGE: Methods with LLMs}}} \\ 
Qwen2.5-7B   & 0.6383 & 0.5118 & 0.9341 & 0.9361 & 0.9351 & 0.7646 & 0.6887 & 0.7247 & 0.5016 & 0.7969 & 0.5447 & 0.8320\\
+SFT & 0.6694 & 0.5532 & 0.9521 & 0.9481 & 0.9501 & 0.7848 & 0.7677 & 0.7762 & 0.6239 & 0.8568 & 0.7233 & 0.8887\\
+MRAG & 0.6774 & 0.5567 & 0.9561 & 0.9574 & 0.9568 & 0.7905 & 0.7641 & 0.7771 & 0.6216 & 0.8448 & 0.7274 & 0.8995 \\
\rowcolor{gray!20} 
+JUSTICE & \textbf{0.7090} &\textbf{ 0.5969 }& \textbf{0.9691} & \textbf{0.9661} &\textbf{ 0.9676} & \textbf{0.7971} & \textbf{0.7900} & \textbf{0.7935} &\textbf{ 0.6371} & \textbf{0.8638} & \textbf{0.7379} & \textbf{0.9116}\\
Baichuan2-7B & 0.6184 & 0.5094 & 0.9172 & 0.9185 & 0.9178 & 0.6893 & 0.7359 & 0.7118 & 0.5157 & 0.8192 & 0.5382 & 0.8407\\
+SFT & 0.6646 & 0.5462 & 0.9431 & 0.9345 & 0.9388 & 0.7735 & 0.7513 & 0.7622 & 0.6403 & 0.8554 & 0.7304 & 0.8717 \\
+MRAG & 0.6676 & 0.5510 & 0.9471 & 0.9464 & 0.9468 & 0.7778 & 0.7539 & 0.7657 & 0.6348 & 0.8559 & 0.7246 & 0.8755 \\
\rowcolor{gray!20} 
+JUSTICE & \textbf{0.6984} & \textbf{0.5834} & \textbf{0.9691} & \textbf{0.9634} & \textbf{0.9662} & \textbf{0.8028} & \textbf{0.7847} &\textbf{ 0.7936} & \textbf{0.6494} & \textbf{0.8668} & \textbf{0.7441} & \textbf{0.9110}\\
\Hline
\multicolumn{10}{l}{\emph{\textcolor{gray!120}{LeCaRDv2-Doc: Traditional CVG Methods}}} \\ 
C3VG(Bart) & - & - & - & - & - & - & - & - & 0.1688 & 0.7052 & - & -\\
$\mathrm{EGG_{free}}$ & - & - & - & - & - & - & - & - & 0.2168 & 0.6752 & - & -\\
EGG & - & - & - & - & - & - & - & - & 0.2271 & 0.6905 & - & -\\
\hline
\multicolumn{13}{l}{\emph{\textcolor{gray!120}{LeCaRDv2-Doc: Traditional LJP Methods}}} \\
TopJudge* & 0.3653 & - & 0.2209 & 0.2460 & 0.2075 & 0.1033 & 0.5000 & 0.1713 & - & - & - & -\\
MPBFN* & 0.3277 & - & 0.2963 & 0.3578 & 0.2964 & 0.1038 & 0.5100 & 0.1725 & - & - & - & - \\
NeurJudge* & 0.3643 & - & 0.5756 & 0.5748 & 0.5660 & 0.0930 & 0.4580 & 0.1546 & - & - & - & - \\
\hline
\multicolumn{13}{l}{\emph{\textcolor{gray!120}{LeCaRDv2-Doc: Methods with LLMs}}} \\
Qwen2.5-7B & 0.5512 & 0.4180 & 0.8087 & 0.8300 & 0.8192 & 0.6502 & 0.6539 & 0.6521 & 0.4810 & 0.7420 & 0.5289 & 0.7671\\
+SFT & 0.5796 & 0.4451 & 0.7890 & 0.8100 & 0.7994 & 0.6811 & 0.6741 & 0.6776 & 0.5253 & 0.8155 & 0.6202 & 0.8368\\
+MRAG & 0.5912 & 0.4783 & 0.8317 & 0.8530 & 0.8422 & \textbf{0.6891} & 0.6662 & 0.6775 & 0.5375 & 0.8095 & 0.6152 & 0.8102 \\
\rowcolor{gray!20}
+JUSTICE & \textbf{0.6492} & \textbf{0.4977} & \textbf{0.8877} & \textbf{0.9090} & \textbf{0.8982} & 0.6734 & \textbf{0.7291} & \textbf{0.7001} & \textbf{0.5513} & \textbf{0.8254} &\textbf{ 0.6247} & \textbf{0.8707}\\
Baichuan2-7B & 0.5690 & 0.3872 & 0.7320 & 0.7520 & 0.7419 & 0.6125 & 0.6136 & 0.6131 & 0.5083 & 0.7836 & 0.5428 & 0.7581\\
+SFT & 0.5844 & 0.4256 & 0.7917 & 0.8150 & 0.8032 & 0.6752 & \textbf{0.7058} & 0.6901 & 0.5210 & 0.8249 & 0.6159 & 0.8610 \\
+MRAG & 0.5873 & 0.4426 & 0.8323 & 0.8620 & 0.8469  & 0.6939 & 0.7006 & 0.6972 & 0.5469 & 0.8230 & 0.6050 & 0.8447 \\
\rowcolor{gray!20}
+JUSTICE & \textbf{0.6075} & \textbf{0.4670} &\textbf{ 0.8837} &\textbf{ 0.9060} & \textbf{0.8947} & \textbf{0.7371} & 0.6818 & \textbf{0.7083} & \textbf{0.5542} & \textbf{0.8270} & \textbf{0.6182} & \textbf{0.8659}\\
\Hline
\end{tabular}}
\caption{Overall performance on dataset JuDGE and LeCaRDv2-Doc, where * indicates only a single charge will be evaluated. “METH” denotes METEOR, “BERTS” for BERTScore, and “Prec” for precision. }
\label{tab:overall results}
\vspace{-5pt}
\end{table*}

\begin{table*}[hbtp]
\setlength{\tabcolsep}{2mm} 
\centering
\resizebox{0.9\textwidth}{!}{
\begin{tabular}{lcccccccccccc}
\Hline 
\multirow{2}{*}{\textbf{Models}} & 
\multicolumn{2}{c}{\textbf{Penalty Acc}} & 
\multicolumn{3}{c}{\textbf{Convicting Acc}} & 
\multicolumn{3}{c}{\textbf{Referencing Acc}} & 
\multicolumn{2}{c}{\textbf{Reasoning Section}} & 
\multicolumn{2}{c}{\textbf{Judgment Section}} \\ 
\cmidrule(lr){2-3} \cmidrule(lr){4-6} \cmidrule(lr){7-9} \cmidrule(lr){10-11} \cmidrule(l){12-13} 
 & Prison & Fine & Recall & Prec & F1 & Recall & Prec & F1 & METH & BERTS & METH & BERTS  \\
\Hline
\multicolumn{10}{l}{\emph{\textcolor{gray!120}{JuDGE: Qwen2.5-7B}}} \\
\rowcolor{gray!20} 
JUSTICE & \textbf{0.7090} &\textbf{ 0.5969 }& \textbf{0.9691} & \textbf{0.9661} &\textbf{ 0.9676} & \textbf{0.7971} & 0.7900 & \textbf{0.7935} &\textbf{ 0.6371} & \textbf{0.8638} & \textbf{0.7379} & \textbf{0.9116}\\
w/o RJER  & 0.6530 & 0.5135 & 0.9301 & 0.9311 & 0.9306 & 0.6336 & 0.7276 & 0.6674 & 0.6044 & 0.8505 & 0.6980 & 0.8783\\
w/o ICE  & 0.5534 & 0.4381 & 0.9541 & 0.9561 & 0.9551 & 0.6902 & 0.6826 & 0.6864 & 0.6261 & 0.8560 & 0.6920 & 0.8937 \\
W/o RJER\&ICE  & 0.6751 & 0.5144 & 0.9371 & 0.9391 & 0.9381 & 0.5032 &\textbf{0.8554} & 0.6336 & 0.5908 & 0.8546 & 0.6941 & 0.8997\\
\hline
\multicolumn{10}{l}{\emph{\textcolor{gray!120}{JuDGE: Baichuan2-7B}}} \\ 
\rowcolor{gray!20} 
JUSTICE & \textbf{0.6984} & \textbf{0.5834} & \textbf{0.9691} & \textbf{0.9634} & \textbf{0.9662} & \textbf{0.8028} & 0.7847 &\textbf{ 0.7936} & \textbf{0.6494} & \textbf{0.8668} & \textbf{0.7441} & \textbf{0.9110}\\
w/o RJER & 0.6878 & 0.4919 & 0.9411 & 0.9415 & 0.9413 & 0.7378 & 0.7694 & 0.7533 & 0.6347 & 0.8662 & 0.7326 & 0.9086 \\
w/o ICE & 0.5485 & 0.4325 & 0.9261 & 0.9265 & 0.9263 & 0.7063 & 0.6847 & 0.6953 & 0.6301 & 0.8597 & 0.6966 & 0.8931 \\
w/o RJER\&ICE & 0.6457 & 0.4568 & 0.9321 & 0.9325 & 0.9323 & 0.5522 & \textbf{0.8229 }& 0.6610 & 0.6036 & 0.8587 & 0.6933 & 0.8927 \\
\Hline
\multicolumn{10}{l}{\emph{\textcolor{gray!120}{LeCaRDv2-Doc: Qwen2.5-7B}}} \\ 
\rowcolor{gray!20}
JUSTICE & \textbf{0.6492} & \textbf{0.4977} & \textbf{0.8877} & \textbf{0.9090} & \textbf{0.8982} & \textbf{0.6734} & 0.7291 & \textbf{0.7001} & \textbf{0.5513} & \textbf{0.8254} &\textbf{ 0.6247} & \textbf{0.8707}\\
w/o RJER & 0.5979 & 0.4283 & 0.7903 & 0.8090 & 0.7996 & 0.6063 & 0.6296 & 0.6177 & 0.5195 & 0.8186 & 0.5866 & 0.8491 \\
w/o ICE & 0.5387 & 0.4536 & 0.8790 & 0.8930 & 0.8859 & 0.6597 & 0.6538 & 0.6567 & 0.5423 & 0.8244 & 0.6035 & 0.8590 \\
w/o RJER\&ICE & 0.5994 & 0.3975 & 0.8237 & 0.8440 & 0.8337 & 0.3725 & \textbf{0.7908} & 0.5065 & 0.4896 & 0.8124 & 0.5955 & 0.8593 \\
\hline
\multicolumn{13}{l}{\emph{\textcolor{gray!120}{LeCaRDv2-Doc: Baichuan2-7B}}} \\
\rowcolor{gray!20}
JUSTICE & \textbf{0.6075} & \textbf{0.4670} &\textbf{ 0.8837} &\textbf{ 0.9060} & \textbf{0.8947} & \textbf{0.7371} & 0.6818 & \textbf{0.7083} & \textbf{0.5542} & \textbf{0.8270} & \textbf{0.6182} & \textbf{0.8659}\\
w/o RJER & 0.5990 & 0.4365 & 0.8017 & 0.8320 & 0.8166 & 0.6186 & 0.6603 & 0.6388 & 0.5319 & 0.8211 & 0.6218 & 0.8588 \\
w/o ICE & 0.5438 & 0.4446 & 0.8703 & 0.8810 & 0.8756 & 0.6754 & 0.6538 & 0.6644 & 0.5468 & 0.8264 & 0.6129 & 0.8609 \\
w/o RJER\&ICE & 0.5498 & 0.3751 & 0.7523 & 0.7790 & 0.7654 & 0.4277 & \textbf{0.7024} & 0.5316 & 0.5066 & 0.8184 & 0.5920 & 0.8489 \\
\Hline
\end{tabular}}
\caption{Ablation Study on dataset JuDGE and LeCaRDv2-Doc.}
\label{tab:ablation results}
\vspace{-10pt}
\end{table*}

\section{Experiment}

In this section, we conduct experiments to address the following research questions:
\begin{itemize}[leftmargin=*]
    \item \textbf{RQ1}: \emph{Does the proposed JUSTICE framework improve the performance of judgment document generation?}
    \item \textbf{RQ2}: \emph{What are the individual contributions of the RJER and ICE modules to the overall performance?}
    \item \textbf{RQ3}: \emph{How effective is the ICE module in improving the accuracy of the final judgment?}
    \item \textbf{RQ4}: \emph{How do the type and quantity of judicial elements from RJER affect the performance of the ICE module?}
\end{itemize}

\subsection{Experiment Setting}

\paragraph{Datasets.}

We evaluate our method on two datasets. Our primary benchmark is JuDGE~\cite{su2025judge}, a large-scale dataset tailored for Chinese judgment document generation. To test out-of-distribution (OOD) performance, we also constructed the LeCaRDv2-Doc test set from the LeCaRDv2~\cite{li2024lecardv2} case retrieval dataset. Further details are provided in Appendix A.

\paragraph{Evaluation Metrics.}

We adopt the multi-dimensional evaluation framework from the JuDGE benchmark. For legal accuracy, we use Precision, Recall, and F1-Score to measure \textbf{Convicting Accuracy} (charge) and \textbf{Referencing Accuracy} (law articles). We also measure the correctness of prison terms and fines using \textbf{Penalty Accuracy}. For textual quality, we use \textbf{METEOR}~\cite{banerjee2005meteor} and \textbf{BERTScore}~\cite{zhang2019bertscore} to assess semantic coherence. Further details are provided in Appendix B.

\paragraph{Baselines.}

We compare our method against three categories of baselines. (1) \textbf{CVG Methods}: \textbf{C3VG}~\cite{yue2021circumstances} enhances court view generation by selecting circumstances; \textbf{EGG}~\cite{yue2024event} grounds generation on extracted events, with an LLM-free variant \bm{$\mathrm{EGG_{free}}$}. (2) \textbf{LJP Methods}: \textbf{TopJudge}~\cite{zhong2018legal} models subtask dependencies topologically; \textbf{MBPFN}~\cite{yang2019legal} uses a multi-perspective bi-feedback network; and \textbf{NeurJudge}~\cite{yue2021neurjudge}, which separates fact representation based on upstream predictions. (3) \textbf{LLM-based Methods}: We include standard \textbf{ICL} and \textbf{SFT}, and \textbf{MRAG}~\cite{su2025judge}, which directly uses knowledge retrieved from statute and case corpora to augment generation.

\subsection{Overall Performance (RQ1)}
As shown in Table~\ref{tab:overall results}, our proposed JUSTICE framework significantly outperforms all baselines on both the in-domain JuDGE and OOD LeCaRDv2-Doc datasets. While traditional methods are ineffective on the small-scale JuDGE training set ($\sim$ 2k samples), JUSTICE's superiority stems from its explicit modeling of the Pre-Judge phase, which enhances both judgmental accuracy and textual quality.

\textbf{JUSTICE excels in legal accuracy by operationalizing the Pre-Judge phase.}
On both datasets, our framework consistently outperforms the strongest MRAG baseline in legal accuracy (e.g., Convicting F1 on JuDGE: 0.9676 vs. 0.9601). This robust performance is attributed to the ICE module, whose verifiable intermediate conclusion provides a more reliable anchor for the final judgment than directly using retrieved knowledge.

\textbf{JUSTICE generates higher quality and more coherent legal text.}
JUSTICE also leads in textual quality, improving BERTScore over all baselines (e.g., 0.9116 vs. 0.8995 on JuDGE). This improvement is attributed to the JUS module synthesizes the final document from a structured conclusion and a formal template, ensuring higher narrative coherence than assembling disparate information.

\subsection{Ablation Study (RQ2)}

To isolate the contributions of our modules, we conduct an ablation study focusing on RJER and ICE. The JUS module is not ablated, as it is the indispensable final generation stage. As shown in Table~\ref{tab:ablation results}, removing either RJER or ICE leads to a significant performance degradation, confirming their complementary and crucial roles within the full JUSTICE framework.

\paragraph{RJER provides the crucial referential foundation.}
Removing RJER results in a significant drop in Referencing Accuracy (e.g., F1 falls from 0.7935 to 0.6674 on JuDGE). This demonstrates the importance of RJER in retrieving a comprehensive set of judicial elements necessary for identifying the correct statutes.

\paragraph{ICE is vital for accurate judgment prediction.}
Removing ICE, which bypasses the Pre-Judge step, leads to a substantial decline in Penalty Accuracy (e.g., prison term score drops from 0.7090 to 0.5534 on JuDGE). This confirms that its structured intermediate conclusion is critical for determining correct sentencing.

\textit{Together, these results confirm the complementary roles of our components: RJER is essential for establishing legal grounding and accurate referencing, while ICE is critical for translating these elements into precise judgment predictions, particularly for sentencing.}

\subsection{Effectiveness of ICE (RQ3)}

\begin{figure}[htbp]
\centering
\includegraphics[width=0.98\linewidth]{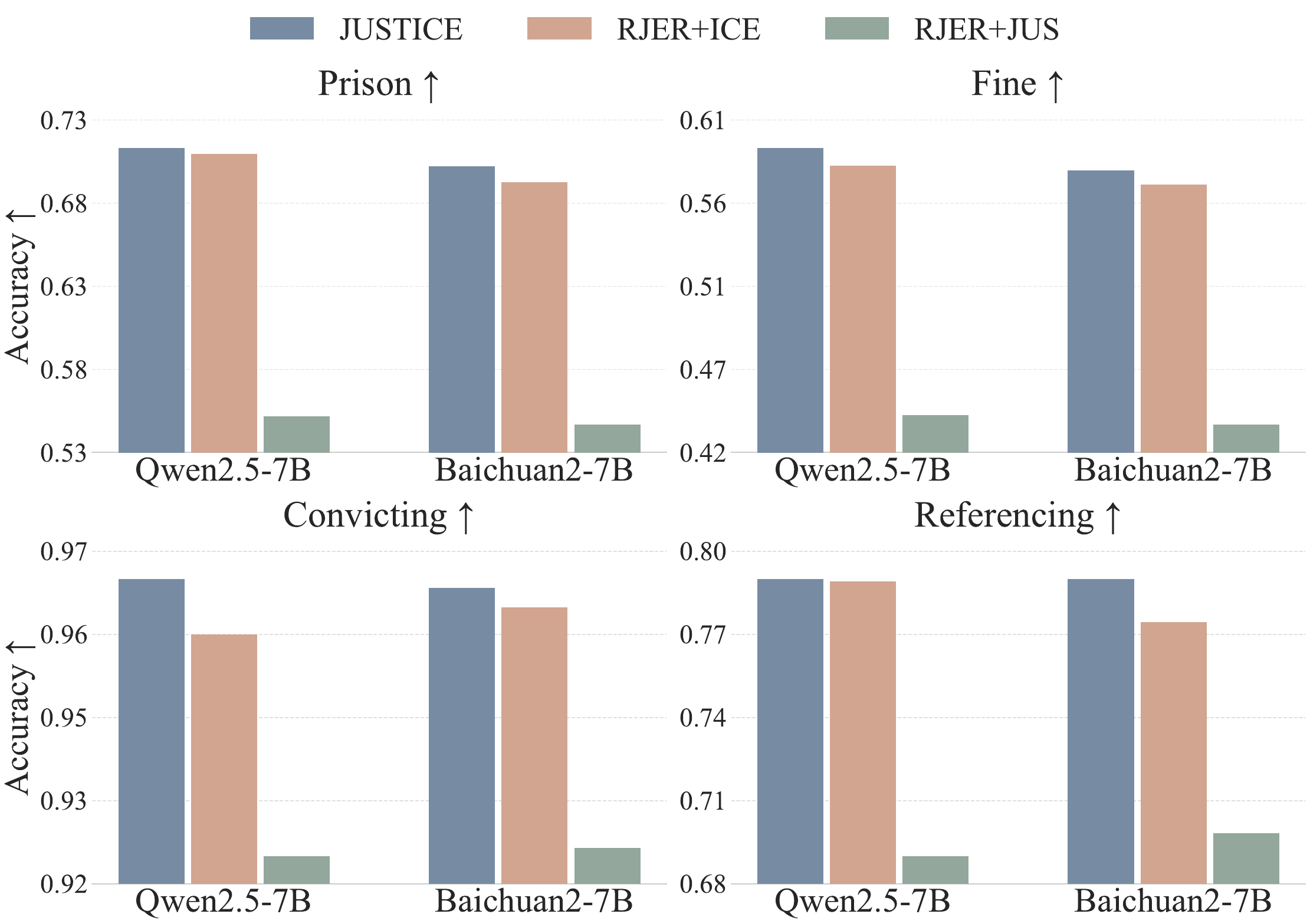} 
\caption{Impact of the ICE module on judgmental accuracy.}
\label{fig:ICE}
\vspace{-10pt}
\end{figure}

To evaluate the effectiveness of the ICE module, we conduct a comparative analysis of the full JUSTICE framework against two ablated variants: RJER+ICE, which concludes its process after the Pre-Judge stage, and RJER+JUS, which bypasses this stage entirely. As illustrated in Figure~\ref{fig:ICE}, the results confirm that ICE is critical for legal accuracy.

\paragraph{Bypassing the Pre-Judge stage harms legal accuracy.}
Removing the ICE module and feeding retrieved elements directly to the writer results in a substantial performance drop (e.g., Convicting F1 plummets from 0.968 to 0.955 with Qwen2.5-7B). This confirms that the intermediate conclusion generated by ICE is an indispensable step for ensuring legal accuracy.

\paragraph{ICE provides a strong foundation for the final writer.}
Figure~\ref{fig:ICE} also shows that the intermediate output from RJER+ICE is already highly accurate, significantly outperforming the direct RJER+JUS variant. This proves that ICE provides a high-quality, structured conclusion, which the full JUSTICE model then builds upon to achieve state-of-the-art performance.

\subsection{Impact of RJER's Output on ICE (RQ4)}

\begin{figure}[htbp]
\centering
\includegraphics[width=0.98\linewidth]{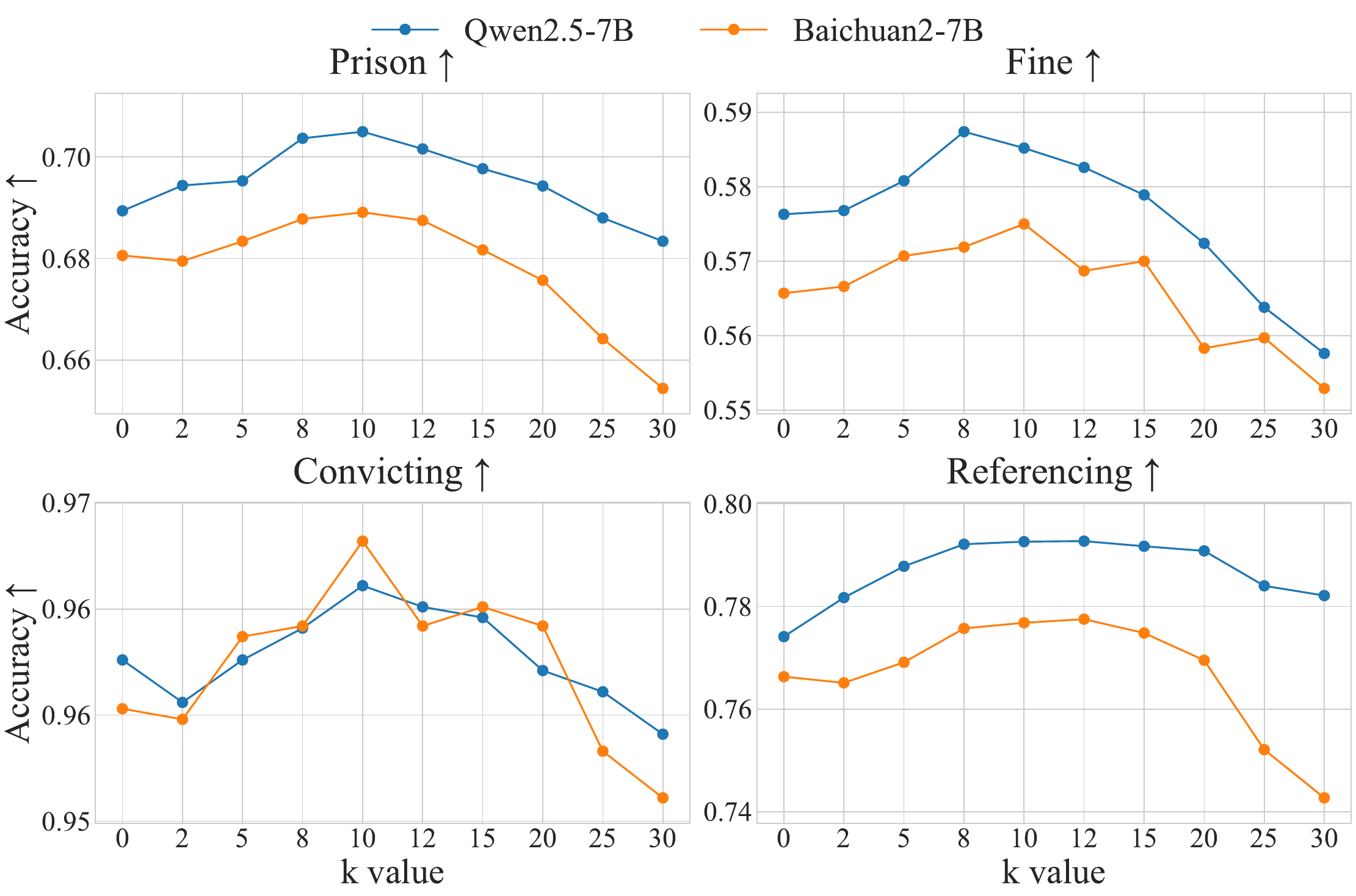} 
\caption{Impact of retrieved article quantity ($k_2$).}
\label{fig:laws_k}
\vspace{-10pt}
\end{figure}

\begin{figure*}[t]
\centering
\includegraphics[width=0.98\textwidth]{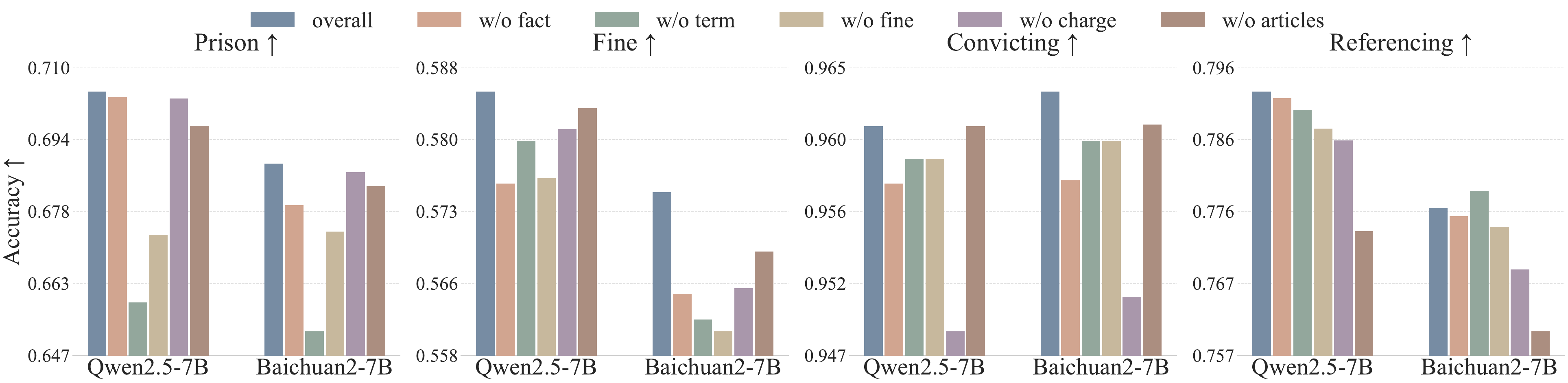} 
\caption{Impact of different reference judicial element types.}
\label{fig:case_info}
\vspace{-5pt}
\end{figure*}

We analyze how the quantity and type of judicial elements from RJER affect the performance of the ICE module, with results shown in Figure~\ref{fig:laws_k} and Figure~\ref{fig:case_info}.

\paragraph{Impact of Element Quantity.}
Figure~\ref{fig:laws_k} illustrates that performance is sensitive to the number of retrieved articles, $k_2$. For most metrics, performance peaks when $k_2$ is around 10 and then declines. This trend suggests that while sufficient context is crucial, an excess of articles can introduce noise, validating the design of our reranking model.

\paragraph{Impact of Element Types.}
Figure~\ref{fig:case_info} shows that each judicial element retrieved by RJER is crucial for the Pre-Judge stage. Ablating specific element types leads to targeted performance degradation. For instance, removing retrieved laws harms Referencing Accuracy, while removing precedent elements such as the charge causes the most significant drop in the corresponding metric. This demonstrates that RJER’s multi-faceted retrieval is essential.

\begin{figure*}[t]
\centering
\includegraphics[width=0.98\textwidth]{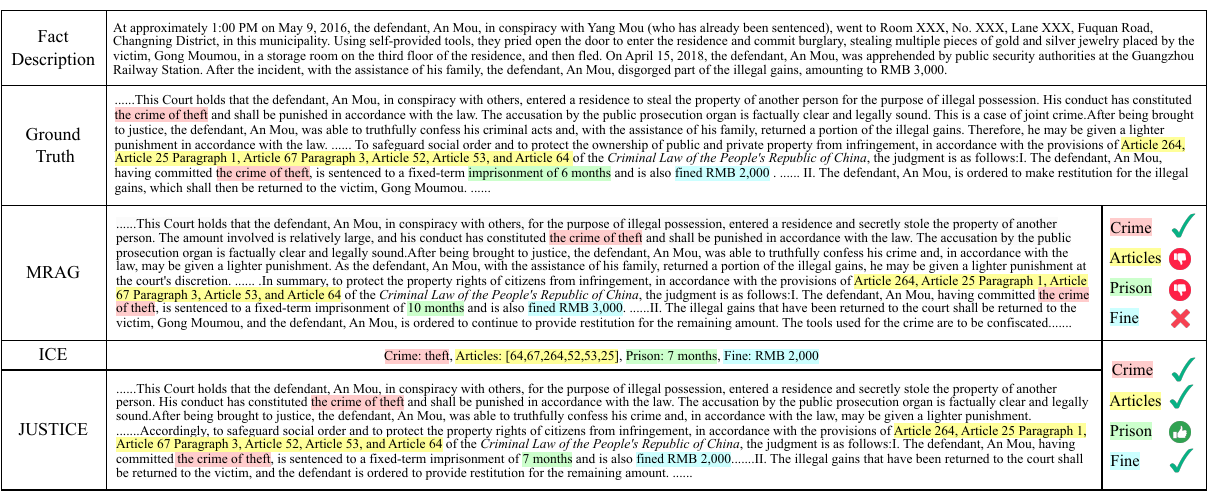} 
\caption{Case Study}
\label{fig:case_study}
\vspace{-10pt}
\end{figure*}

\subsection{Case Study}

To provide an intuitive understanding of our framework's advantages, we present a case study in Figure~\ref{fig:case_study} comparing the outputs of our JUSTICE framework with the MRAG baseline for a theft case.

\paragraph{Limitations of Direct Generation.}
The MRAG baseline exhibits notable deficiencies. While it correctly identifies the ``crime of theft," it fails to construct a complete legal basis and makes incorrect predictions for both the prison term and fine, demonstrating the limitations of direct generation.

\paragraph{Accuracy Ensured by Intermediate Conclusion.}
In stark contrast, our JUSTICE framework demonstrates exceptional precision. The key advantage stems from the ICE module, which generates a highly accurate intermediate conclusion, correctly identifying the crime, fine amount, and all required law articles. The final JUS module then synthesizes this high-quality input, resulting in a vastly superior and legally sound judgment document.

\section{Literature Review}

\subsection{Legal Judgment Prediction}

Legal Judgment Prediction (LJP) seeks to automatically forecast case outcomes from fact descriptions. Foundational work centered on modeling dependencies between sub-tasks like charge and article prediction~\cite{zhong2018legal,yang2019legal}. To handle real-world complexities, recent research has shifted towards multi-task and multi-label settings~\cite{li2025legal}, and has explored incorporating structured information like events~\cite{feng2022legal} or numerical evidence~\cite{gan2023exploiting} to tackle confusing cases. A significant advancement has been the decomposition of the prediction process. For instance, NeurJudge~\cite{yue2021neurjudge} used upstream predictions to separate fact representations, while later works such as RLJP~\cite{wu2022towards} and ADAPT~\cite{deng2024enabling} introduced explicit reasoning steps, generating textual rationales or following an "Ask-Discriminate-Predict" path to guide the model's final prediction.

This trend towards decomposed reasoning in LJP inspires our work. While LJP focuses on prediction, we adapt its core principle, positing that emulating a judge's preliminary conclusion is a crucial prerequisite for the more complex task of judgment generation.

\subsection{Judgment Document Generation}

The proliferation of LLMs has broadened the research scope from LJP to full-text Judgment Document Generation. Initial efforts concentrated on generating specific sections of a judgment, most notably the court's reasoning (a task known as Court View Generation, or CVG), with methods like C3VG~\cite{yue2021circumstances}, EGG~\cite{yue2024event}, and the concept-guided LeGen~\cite{xu2024divide}. The dominant paradigm has now shifted to Retrieval-Augmented Generation (RAG), where state-of-the-art models like MRAG~\cite{su2025judge} and the precedent-enhanced PLJP~\cite{wu2023precedent} leverage retrieved statutes and precedent cases to directly augment the final text generation.

However, these RAG-based methods lack the crucial intermediate reasoning stage of human judges. Our JUSTICE framework is designed to bridge this exact gap by explicitly modeling this Pre-Judge process, ensuring the final document is grounded in a verifiable intermediate decision.

\section{Conclusion}
In this paper, we propose JUSTICE, a framework that improves legal coherence and accuracy in automated judgment generation by explicitly modeling the Pre-Judge phase omitted by prior work. By integrating three dedicated components—RJER, a referential element retriever; ICE, an intermediate conclusion emulator; and JUS, a unified document synthesizer—our method explicitly models the "Search → Pre-Judge → Write" cognitive workflow. Our findings emphasize the importance of this modeling approach and highlight its potential for enhancing other complex generation tasks within the legal domain.
\FloatBarrier

\bibliography{chapter_wbl/ref}

\appendix

\section{LeCaRDv2-Doc Dataset }

To further evaluate the out-of-distribution (OOD) generalization capabilities of the models, we introduce the LeCaRDv2-Doc dataset, which is processed from the original LeCaRDv2 corpus.  The basic statistics of this dataset are summarized in Table~\ref{tab:lecardv2-doc}.

Unlike the JuDGE dataset, which is partitioned into a standard training and test set from the same distribution, LeCaRDv2-Doc is specifically structured to assess model robustness against distributional shifts.  We randomly selected 500 fact-judgment pairs to serve as a dedicated OOD test set.  The remaining 10,039 documents from LeCaRDv2 constitute the case retrieval corpus, which can be used by retrieval-augmented models to find relevant cases for the OOD test instances.

The characteristics of the LeCaRDv2-Doc test set present a different profile compared to the JuDGE dataset.  As shown in the table, the 500 test cases cover 56 unique charges and 95 unique criminal law provisions.  The average fact length is considerably shorter at approximately 352 characters, compared to over 650 in JuDGE.  This suggests that the cases might be more concisely described.  Interestingly, despite the shorter fact descriptions, each document references an average of 5.01 statutory articles, which is higher than the 4.31 articles per document in JuDGE.  This implies that the cases, while factually less verbose, may involve a comparable or even greater degree of legal complexity, making them suitable for challenging the models' legal reasoning and article citation abilities in an OOD context.

\begin{table}[htbp]
  \centering 
  \resizebox{\columnwidth}{!}{
  \begin{tabular}{lr}
    \toprule
    \textbf{Statistic} & \textbf{Number} \\
    \midrule
    Total Fact-Judgment Pairs & 500 \\
    Unique Charges & 56 \\
    Unique Criminal Law Provisions & 95 \\
    Avg. Fact Length & 351.95 \\
    Avg. Reasoning Length & 283.49 \\
    Avg. Judgment Result Length & 174.50 \\
    Avg. Full Document Length & 1,571.78 \\
    Avg. Charges per Document & 1.11 \\
    Avg. Statutory Articles per Document & 5.01 \\
    \midrule
    External Judgment Documents Corpus Size & 10,039 \\
    \bottomrule
  \end{tabular}%
  }
    \caption{Basic Statistics of the LeCaRDv2-Doc Dataset.}
  \label{tab:lecardv2-doc}
\end{table}

\section{Evaluation Details}
We adopt a four-part framework to evaluate the quality of generated judgment documents, assessing critical aspects of legal integrity.

\subsection{Penalty Accuracy}
This metric assesses the precision of penalties, such as prison terms and fines. Given their real-world impact, accuracy is crucial. The score, $S_{penalty}$, quantifies the deviation from the ground-truth values using a normalized difference.
\begin{equation}
    S_{penalty} = 1 - \frac{|V_{gen} - V_{gt}|}{\max(V_{gen}, V_{gt})}
\end{equation}
Here, $V_{gen}$ is the generated value and $V_{gt}$ is the ground-truth value.

\subsection{Convicting Accuracy}
This evaluates the model's ability to correctly identify all charges for which a defendant is convicted. Performance is measured using standard classification metrics:
\begin{itemize}
    \item \textbf{Precision:} The proportion of predicted charges that are correct.
    \item \textbf{Recall:} The proportion of actual charges that were correctly identified.
    \item \textbf{F1 Score:} The harmonic mean of precision and recall, providing a balanced measure.
\end{itemize}

\subsection{Referencing Accuracy}
This metric assesses the accuracy of statutory articles cited in the judgment. It penalizes both the omission of relevant statutes and the inclusion of irrelevant ones. It is evaluated with:
\begin{itemize}
    \item \textbf{Precision:} The fraction of cited statutes that are relevant.
    \item \textbf{Recall:} The fraction of relevant statutes that are correctly cited.
    \item \textbf{F1 Score:} A consolidated score balancing completeness and correctness.
\end{itemize}

\subsection{Documenting Similarity to Ground Truth}
This measures the semantic faithfulness of the \textit{Judicial Reasoning} and \textit{Judgment Result} sections. It focuses on meaning rather than lexical overlap, as traditional metrics often fail to capture legal nuances. The metrics used are:
\begin{itemize}
    \item \textbf{METEOR:} An alignment-based metric that handles synonyms and paraphrasing.
    \item \textbf{BERTScore:} Leverages contextual embeddings to compare deep semantic similarity.
\end{itemize}

\section{Potential for Human-in-the-Loop Correction}
To evaluate the upper-bound performance and the potential for human-in-the-loop intervention, this supplementary experiment compares the standard JUSTICE framework with a variant JUSTICE+GT. In the JUSTICE+GT setting, the intermediate conclusion generated by the ICE module is replaced with the ground-truth (GT) tuple, providing an ideal input for the final JUS synthesizer.

\begin{figure}[htbp]
\centering
\includegraphics[width=0.98\linewidth]{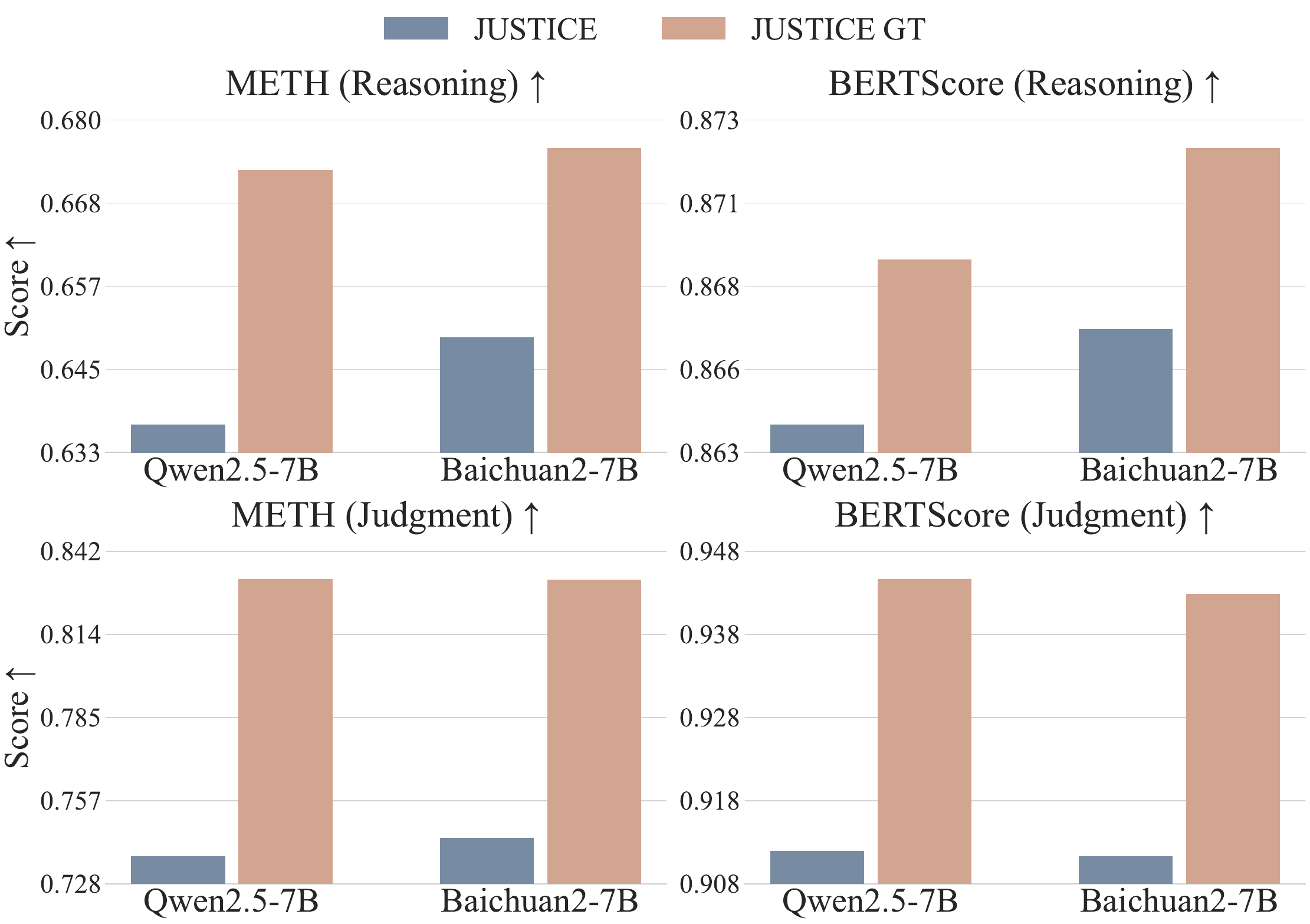} 
\caption{ Upper-Bound Performance with a Perfect Intermediate Conclusion. }
\label{fig:GT}
\end{figure}

As illustrated in Figure~\ref{fig:GT}, the results yield two key insights. First, the standard JUSTICE framework exhibits robust performance, establishing a strong baseline. Second, the JUSTICE+GT variant demonstrates a significant performance gain, particularly in textual quality metrics like BERTScore, which indicates that the framework's performance ceiling is substantially higher. This underscores a critical advantage of our modular design: the intermediate conclusion from ICE serves as an effective checkpoint. The possibility of refining this stage, either through future model improvements or human expert correction, is crucial for enhancing the reliability and trustworthiness of automated systems in high-stakes domains like the legal field.

\begin{figure*}[t]
    \centering
    \includegraphics[width=0.98\linewidth]{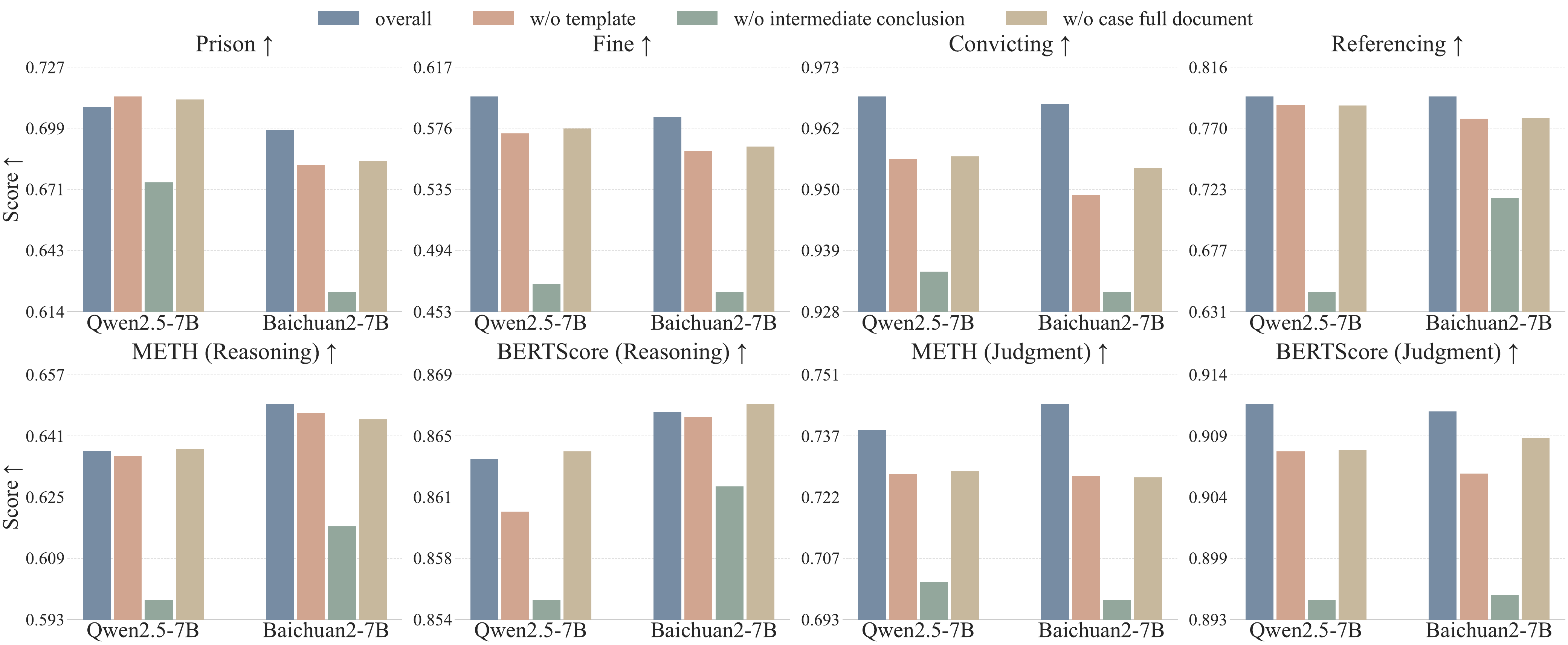}
    \caption{Impact of Information Sources on the JUS Module.}
    \label{fig:doc_info}
\end{figure*}

\section{Impact of Information on Final Synthesis}

To assess the contribution of each information source to the final document synthesis in the JUS module, we conducted an ablation study. We systematically removed one input source at a time—the intermediate conclusion from ICE, the full precedent case document, and the document template—to observe the impact on both legal accuracy and textual quality. The results are illustrated in Figure~\ref{fig:doc_info}.

The analysis reveals that each component is critical for generating a high-quality judgment. Removing the intermediate conclusion leads to the most substantial performance degradation across nearly all metrics, particularly in legal accuracy like Convicting and Penalty Accuracy. This underscores its role as the indispensable logical foundation for the final document. The absence of the full precedent case document, which serves as a stylistic and structural exemplar, results in a noticeable drop in textual quality scores (METEOR and BERTScore). Similarly, omitting the document template harms the structural integrity and coherence of the generated text. These findings validate our design choice to ground the JUS module on this multi-faceted set of inputs, ensuring the final judgment is legally sound, stylistically appropriate, and structurally correct.

\section{Limitations}
Our framework is designed for the Chinese legal system.  Its applicability to other jurisdictions has not yet been validated. Practical deployment in judicial settings also necessitates strict offline processing for data confidentiality.  This constraint precludes the use of powerful, state-of-the-art LLMs, which are predominantly accessed via APIs, and thus restricts the system to smaller, locally deployed LLMs.  The quality of the generated document is also influenced by the single precedent case provided by the retrieval module.  This dependency highlights the importance of a robust case retrieval model to ensure the quality of the generated output.  For future work, we plan to adapt the framework to other legal systems and investigate techniques to improve the system's robustness with respect to the retrieved precedents.

\section{Ethics Statement}
Our framework is developed as an assistive tool for legal professionals, intended to support the judicial workflow by automating preliminary document drafting rather than replacing human judges. Accountability for all legal decisions remains entirely with qualified judicial personnel, who must review and validate any AI-generated output. For any practical deployment, a thorough analysis of potential societal biases is necessary, and robust human oversight mechanisms must be established.

\section{Implement Details}

All experiments were conducted on a server equipped with eight NVIDIA RTX 4090 GPUs. We selected two large language models for our experiments: Qwen2.5-7B-Instruct and Baichuan2-7B-Chat. For the generation phase, we set the temperature to 0.1, the top-k sampling parameter to 1, and the maximum number of new tokens to 3000 to ensure deterministic and consistent outputs.

For the RJER module, we set the number of initial retrieved articles $k_1$ to 100 and the number of reranked articles $k_2$ to 10. For model training, we employed full-parameter fine-tuning using the LLaMA-Factory framework. To manage memory usage across multiple GPUs, we utilized the DeepSpeed ZeRO-3 optimization with CPU offload. The models were trained for 10 epochs with a learning rate of 2e-5. We used a per-device training batch size of 1 and set the number of gradient accumulation steps to 8, resulting in an effective batch size of 8.

\section{The prompt of ICE}
\begin{figure*}[htbp]
    \centering
    \includegraphics[width=0.98\linewidth]{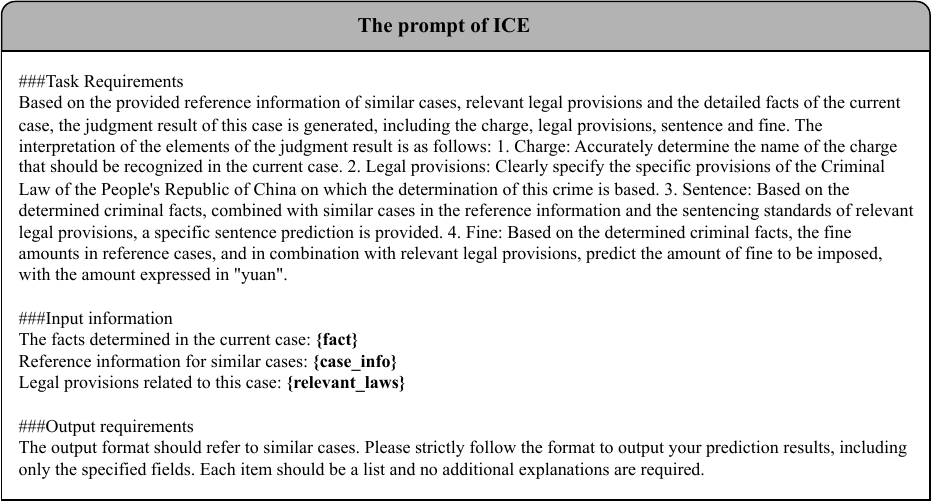}
    \caption{Prompt for the ICE Module.}
    \label{fig:prompt_ice}
\end{figure*}

As shown in Figure~\ref{fig:prompt_ice}, the prompt for the Intermediate Conclusion Emulator (ICE) is designed to guide the LLM in generating a structured pre-judgment conclusion. It provides the model with the facts of the current case, the key elements extracted from a similar precedent case, and a list of relevant legal articles. The prompt explicitly instructs the model to output a tuple containing the predicted charge, applicable laws, prison sentence, and fine, thereby creating a verifiable checkpoint before the final document synthesis.

\section{The prompt of JUS}

\begin{figure*}[htbp]
    \centering
    \includegraphics[width=0.98\linewidth]{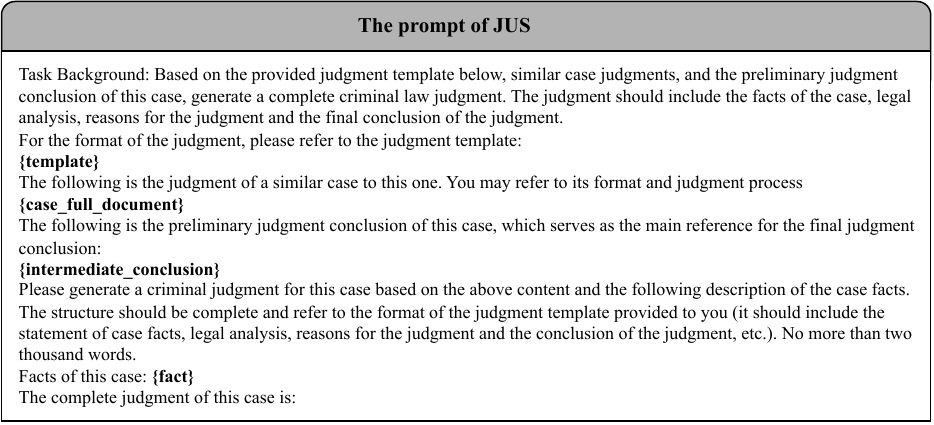}
    \caption{Prompt for the JUS Module.}
    \label{fig:prompt_jus}
\end{figure*}

The prompt for the Judicial Unified Synthesizer (JUS), as shown in Figure~\ref{fig:prompt_jus}, is engineered to guide the LLM in crafting the final, complete judgment document. It integrates multiple sources of information: the original case facts, the structured intermediate conclusion from the ICE module, the full text of a similar precedent case for stylistic reference, and a formal document template to ensure structural consistency. This comprehensive prompt directs the model to synthesize these elements into a legally coherent and well-formatted final judgment.

\end{document}